\documentclass{article}

\usepackage{colm2024_conference}
\usepackage{booktabs}
\usepackage{float}
\usepackage{amsmath}
\usepackage{amsfonts}
\usepackage{nicefrac}
\usepackage{enumitem}
\usepackage{graphicx}
\graphicspath{{fig/}}
\usepackage{algorithm}
\usepackage{algpseudocode}
\usepackage{xcolor}
\hypersetup{
  hypertexnames=false,
  pdftitle={H-Scale: Hessian-Guided Scale Refinement for NVFP4 Sub-Byte LLM Inference},
  pdfauthor={Hao Yu, Zheng Li, Jianwei Zhang, Dayiheng Liu},
  pdfsubject={Technical Report}
}

\title{H-Scale: Hessian-Guided Scale Refinement for NVFP4 Sub-Byte LLM Inference}

\author{%
\textbf{Hao Yu} \quad
\textbf{Zheng Li} \quad
\textbf{Dayiheng Liu} \quad \textbf{Jianwei Zhang}\thanks{Corresponding author.} 
\\
\vspace{1.5mm}
Qwen Team, Alibaba Inc.
}

\begin{document}

\maketitle

\begin{abstract}
  The NVIDIA Blackwell architecture, with native support for the ultra-fine-grained NVFP4 format, opens new opportunities for accelerating large language model (LLM) inference. NVFP4's micro-block design, such as a group size of 16, offers strong representational flexibility for capturing local weight distributions and isolating outliers, but it also introduces a large and highly sensitive space of per-group scaling factors. Existing post-training quantization (PTQ) methods primarily focus on refining quantized weight values, leaving this scale-selection step underexplored. To address this gap, we propose \textbf{H-Scale}, a lightweight post-processing method for NVFP4 per-group scale refinement. Instead of minimizing plain weight reconstruction error, H-Scale selects hardware-valid group scales using a diagonal second-order proxy derived from calibration activations, thereby targeting layer output perturbation more directly. It is designed as a drop-in replacement for RTN-style scale selection in diverse NVFP4 pipelines, requires only modest offline calibration, and introduces strictly zero overhead at inference time. Under a fixed evaluation protocol, experiments on mainstream LLMs show that H-Scale generally improves a broad range of NVFP4 baselines and brings several variants closer to the BF16 reference.
  \end{abstract}

  \section{Introduction}

  Large language models (LLMs) have demonstrated strong capabilities across diverse natural language tasks~\citep{brown2020language,touvron2023llama}. Yet their large parameter counts impose severe demands on memory bandwidth and inference throughput~\citep{pope2022efficientlyscalingtransformerinference}. Post-training quantization (PTQ) has therefore become an important technique for compressing high-precision model tensors (BF16/FP16) into lower-bit formats that are faster and cheaper to execute~\citep{gholami2021surveyquantizationmethodsefficient}. The field is also undergoing a hardware-software co-design shift from integer-based formats (INT8/INT4) toward 4-bit floating-point (FP4) representations~\citep{rouhani2023microscalingdataformatsdeep}. 
  
  The NVIDIA Blackwell architecture introduces native tensor core acceleration for the \textbf{NVFP4} format~\citep{nvidia2024blackwell}. NVFP4 typically uses an ultra-fine-grained structure: weights are organized into micro-blocks (e.g., $g=16$), where an E2M1 4-bit value grid is coupled with a shared E4M3 8-bit scale per block and a tensor-level FP32 global scale~\citep{panferov2026quartetiiaccuratellm}. This fine granularity can better capture local weight distributions and isolate outliers, yet it also creates a large space of per-group scales whose calibration strongly affects final accuracy. In NVFP4, each 16-value group introduces its own scale, so scale miscalibration can become a major source of error even when the FP4 value grid is fixed. This challenge is not fully addressed by existing PTQ methods, and recent studies suggest that techniques which transfer well to earlier quantization settings, such as Hadamard-based preprocessing, become less effective under the NVFP4 microscaling regime~\citep{egiazarian2025bridginggappromiseperformance}. 
  
  Advanced PTQ methods such as GPTQ~\citep{frantar2022gptq} and SpinQuant~\citep{liu2024spinquant} reduce quantization error by reconstructing or regularizing low-bit weights. However, these methods remain largely \textit{weight-centric}: they primarily compensate error by modifying the quantized weight values themselves. We argue that in the NVFP4 setting, the large space of \textit{scaling factors} is a comparably important and still underexplored calibration target. To this end, we introduce \textbf{H-Scale}, a targeted refinement method for per-group scale selection in fine-grained formats. Instead of optimizing the discrete weight representation directly, H-Scale uses a diagonal second-order proxy computed from calibration activations to choose group scales. By shifting the scale-selection objective from plain weight reconstruction error to output-aware weighted reconstruction, H-Scale reduces the rounding noise induced by the coarse E2M1 grid more effectively.
  
  Designed as a lightweight post-processing algorithm, H-Scale can be seamlessly grafted onto existing NVFP4 baselines. It runs in minutes and, crucially, introduces \textbf{zero additional cost} at inference time. Our main contributions are threefold:
  \begin{enumerate}
      \item We identify per-group scale selection as an important bottleneck for NVFP4 fidelity and propose H-Scale, a diagonal-Hessian-weighted scale refinement method for ultra-fine-grained quantization.
      \item We formulate a drop-in scale-selection step, i.e., for each group, it enumerates a short window of neighboring FP8 local scales around the baseline choice and keeps the one that minimizes Hessian-weighted reconstruction error. The quantized weights remain in native NVFP4, so inference cost is unchanged.
      \item Through comprehensive evaluations on state-of-the-art LLMs, we report that H-Scale generally yields higher average performance over diverse baseline algorithms. On complex reasoning benchmarks, H-Scale reduces the average gap between NVFP4 and full-precision BF16 models in several settings.
  \end{enumerate}

  \section{Related Work}
  \label{sec:related_work}
  
  \paragraph{Post-Training Quantization (PTQ) for LLMs.} 
PTQ remains the dominant paradigm for LLM compression because of its computational efficiency. The push toward 4-bit precision exposed the limits of naive rounding and motivated stronger error-compensation methods. GPTQ~\citep{frantar2022gptq} reconstructs weights iteratively using second-order information, AWQ~\citep{lin2023awq} protects salient channels through activation-aware scaling, and GPTAQ~\citep{li2025gptaq} further mitigates errors from outlier activations. Orthogonal transform-based methods such as SpinQuant~\citep{liu2024spinquant} instead learn rotations to smooth weight distributions before quantization. These methods established the viability of sub-8-bit quantization, but they mainly improve low-bit fidelity through weight reconstruction, weight reparameterization, or activation-aware protection. H-Scale instead chooses the per-group scales of micro-block formats with a diagonal second-order proxy, leaving the quantized weight codes to the original rounding rule.
  
\paragraph{Hardware-Co-Designed and Fine-Grained Quantization.}
Native 4-bit floating-point arithmetic, as in NVIDIA Blackwell, has catalyzed NVFP4-specific algorithms, where the fine-grained group size ($g=16$) offers high representational capacity but increases scale sensitivity~\citep{panferov2026quartetiiaccuratellm}. MR-GPTQ~\citep{egiazarian2025bridginggappromiseperformance} adapts GPTQ to the microscaling NVFP4 layout and provides a strong baseline. 4over6~\citep{cook2026six} regularizes value distributions through adaptive sub-block formatting, while ArcQuant~\citep{meng2026arcquant} preserves information with residual activation channels. FAAR~\citep{li2026faar} is closely related but adapts the \emph{rounding} behavior to the non-uniform NVFP4 grid. H-Scale is complementary because it keeps the hardware-native rounding rule fixed and optimizes the \emph{per-group scales}, serving as a \textbf{baseline-agnostic refinement for NVFP4 pipelines} and a drop-in enhancement for Blackwell-generation inference engines.
  
  \paragraph{Scale Refinement for LLM Quantization.}
  The critical role of scaling factors in low-bit quantization has motivated a line of work on \textit{refining} or \textit{optimizing} scales beyond naive min-max or per-group max normalization. LSQ~\citep{esser2020learned} learns quantization scales via backpropagation during fine-tuning, but its training cost is prohibitive for modern LLMs. SmoothQuant~\citep{xiao2023smoothquant} migrates activation outliers into weights via layer-wise scale migration to ease symmetric quantization. It focuses on scale placement between activations and weights rather than post-quantization scale tuning. Similar to our approach, SignRoundV2~\citep{signroundv2scale} performs lightweight search over quantization scales before fine-tuning the weights. MR-GPTQ~\citep{egiazarian2025bridginggappromiseperformance} likewise refines scales prior to applying GPTQ in the NVFP4 setting.
  
  H-Scale aligns with this scale-refinement philosophy but differs in two aspects: (1) the objective is a diagonal second-order, output-aware reconstruction loss, and (2) the intervention sits at the final per-group scale-selection step of the NVFP4 pipeline, so the method is complementary to prior reconstruction or structural quantization.
  
  \section{Method}
  \label{sec:method}
  
  In this section, we formulate NVFP4 quantization and introduce H-Scale as a per-group scale refinement method. We first replace plain weight MSE with a diagonal-Hessian weighted reconstruction objective, then choose each group's scale by enumerating a short window of neighboring E4M3-representable local scales.
  \subsection{Preliminaries: NVFP4 Quantization}
  The NVFP4 format optimizes the trade-off between hardware efficiency and expressivity via a micro-block design. We denote the \textit{original} full-precision weight matrix by $\mathbf{W} \in \mathbb{R}^{m \times n}$ and the \textit{quantized} weight matrix by $\hat{\mathbf{W}}$. Each row of $\mathbf{W}$ is partitioned into contiguous groups of size $g=16$ along the input-channel dimension. We write $\mathbf{w}_{r,i} \in \mathbb{R}^{g}$ for the $i$-th group in row $r$, and $\hat{\mathbf{w}}_{r,i}$ for its quantized counterpart.
  
  In hardware, NVFP4 represents each group using a 4-bit E2M1 value grid together with a per-group FP8 E4M3 local scale and a tensor-level FP32 global scale.   We write $\ell_{r,i}$ for the local scale and $s_{\mathrm{g}}$ for the global scale, and define the \emph{effective group scale}
  \begin{equation}
      s_{r,i}
      =
      s_{\mathrm{g}}\,\ell_{r,i}
      \label{eq:effective_scale}
  \end{equation}
  as the scalar actually used to dequantize the group. Formally, the standard quantization function for a group $\mathbf{w}_{r,i}$ is:
  \begin{equation}
      \hat{\mathbf{w}}_{r,i}
      =
      s_{r,i} \cdot \text{Clip}\left(\text{Round}\left(\frac{\mathbf{w}_{r,i}}{s_{r,i}}\right)\right),
  \end{equation}
  where $\text{Round}(\cdot)$ maps values to the nearest FP4 discrete levels and $\text{Clip}(\cdot)$ bounds the result within the FP4 representable range $[-6.0, 6.0]$. Conventionally, $s_{r,i}$ is initialized from the maximum absolute value within the group and then snapped to the hardware-valid scale representation. 
  
  \subsection{Hessian-Guided Scale Objective}
  
  We start from the \emph{layer output reconstruction} objective rather than plain weight MSE. Let $\mathbf{W} \in \mathbb{R}^{m \times n}$ be a weight matrix, $\hat{\mathbf{W}}$ its quantized counterpart, and $\Delta \mathbf{W} = \mathbf{W} - \hat{\mathbf{W}}$. Given a calibration activation matrix $\mathbf{X} \in \mathbb{R}^{N \times n}$, the output mismatch over the calibration set is
  \begin{equation}
      \|\mathbf{X}\mathbf{W}^{\top} - \mathbf{X}\hat{\mathbf{W}}^{\top}\|_F^2
      = \|\mathbf{X}\Delta \mathbf{W}^{\top}\|_F^2
      = \mathrm{Tr}\!\left(\Delta \mathbf{W}\,\mathbf{X}^{\top}\mathbf{X}\,\Delta \mathbf{W}^{\top}\right).
      \label{eq:layer_output_objective}
  \end{equation}
  This objective involves the full second-order weighting matrix $\mathbf{G} = \mathbf{X}^{\top}\mathbf{X}$, which is the Hessian of the squared output reconstruction loss. Directly optimizing it over all NVFP4 group scales would couple the scale choices across input-channel groups through the off-diagonal entries of $\mathbf{G}$, making the exact discrete search combinatorial. We therefore use a diagonal-Hessian surrogate by retaining only
  \begin{equation}
      \mathbf{h} = \mathrm{diag}(\mathbf{X}^{\top}\mathbf{X}) \in \mathbb{R}^{n},
  \end{equation}
  where each $h_j$ measures the activation energy of input channel $j$ on the calibration set. Under this diagonal approximation, the objective decouples across output rows and input-channel groups, yielding a weighted reconstruction loss:
  \begin{equation}
      \mathrm{Tr}\!\left(\Delta \mathbf{W}\,\mathbf{X}^{\top}\mathbf{X}\,\Delta \mathbf{W}^{\top}\right)
      \approx
      \sum_{r=1}^{m}\sum_{j=1}^{n} h_j\,(\Delta W_{rj})^2.
  \end{equation}
  The approximation drops the cross-channel correlation terms. High-energy input channels are still penalized more heavily, because they contribute more strongly to $\mathbf{X}\Delta\mathbf{W}^{\top}$, while weakly activated channels have less influence on the layer output. The scale search therefore spends the limited FP4 resolution on channels that dominate the calibration output error.
  
  For NVFP4, the diagonal Hessian proxy must respect the same group partition. Let $\mathbf{h}_{i} \in \mathbb{R}^{g}$ be the slice of $\mathbf{h}$ aligned with group index $i$. For each output row $r$ and group $i$, we select the effective group scale from the hardware-valid candidate set by minimizing the corresponding weighted reconstruction error:
  \begin{equation}
      \min_{s_{r,i} \in \mathcal{S}_{r,i}}
      \left\|
      \left(\mathbf{w}_{r,i} - \hat{\mathbf{w}}_{r,i}(s_{r,i})\right)
      \odot \sqrt{\mathbf{h}_{i}}
      \right\|_2^2,
      \label{eq:scale_objective}
  \end{equation}
  where $\odot$ denotes element-wise multiplication and $\mathcal{S}_{r,i}$ denotes the finite set of hardware-valid \emph{effective} scales induced by a length-$K{=}16$ bidirectional walk on the E4M3 local-scale grid, constructed in the next subsection. In practice, we compute $\mathbf{h}$ once per layer from the calibration activations and slice it according to the same input-channel groups used by NVFP4. Eq.~\eqref{eq:scale_objective} is a group-wise diagonal surrogate of Eq.~\eqref{eq:layer_output_objective}: it keeps the per-channel Hessian weights and drops cross-channel terms. Given $\mathcal{S}_{r,i}$, we evaluate every candidate and keep the minimizer, and the search is exhaustive over those $K$ scales. When the group weights are replaced by the all-ones vector $\mathbf{h}_{i}=\mathbf{1}$, Eq.~\eqref{eq:scale_objective} degenerates to the plain weight-centric search objective used in our ablations.
  
  \subsection{Efficient Scale Optimization Algorithm}
  \label{sec:scale_opt}
  Optimizing Eq.~\eqref{eq:scale_objective} is difficult because both the FP4 projection and the scale representation are discrete. Our implementation therefore uses a deterministic \emph{one-dimensional scale search} that preserves the original NVFP4 quantization pipeline, i.e., we keep the baseline grouping and value grid unchanged, and only refine the per-group scale.

  \paragraph{Initialization from the baseline quantizer.}
  For each group $(r,i)$, we first obtain an initial effective scale $s_{r,i}^{\mathrm{init}}$ from the underlying quantizer. In the plain RTN variant, this is the usual max-abs scale induced by the group maximum together with the FP8 scale quantization step. H-Scale does not alter this initialization rule and only refines the resulting scale.

  \paragraph{Hardware-valid scale candidate set.}
  Starting from $s_{r,i}^{\mathrm{init}}$, H-Scale evaluates that scale and a few neighboring E4M3-representable scales on both sides.
  The two directions help for different reasons, because E2M1 is a non-uniform grid on $[-6,6]$: a smaller scale stretches $\mathbf{w}/s$ toward $\pm 6$, so outliers may clip while the remaining weights occupy more of the grid. A larger scale compresses $\mathbf{w}/s$ toward zero, so the group drops the coarsest codes $\{\pm 4,\pm 6\}$ and large-magnitude weights are rounded on denser mid-range codes.
  Which side wins is decided by the Hessian-weighted error in Eq.~\eqref{eq:scale_objective}, not by fitting the group range.

  Because $s_{r,i}=s_{\mathrm{g}}\,\ell_{r,i}$ as in Eq.~\eqref{eq:effective_scale}, it is enough to walk on the E4M3 local-scale grid and multiply back by the shared FP32 global scale. Let $Q_{\mathrm{fp8}}(\cdot)$ denote FP8 E4M3 quantization. The initialized local scale is
  \begin{equation}
     \ell_{r,i}^{\mathrm{init}}
     =
     Q_{\mathrm{fp8}}\!\left(\frac{s_{r,i}^{\mathrm{init}}}{s_{\mathrm{g}}}\right),
     \label{eq:fp8_local_scale}
  \end{equation}
  with corresponding effective scale $s_{\mathrm{g}}\,\ell_{r,i}^{\mathrm{init}}$. Thus only $\ell_{r,i}$ is constrained to the hardware FP8 grid, while $s_{\mathrm{g}}$ remains in FP32.

  Let
  \begin{equation}
      \mathcal{L}^{+}=(\lambda_1<\lambda_2<\cdots<\lambda_M)
      \label{eq:e4m3_ladder}
  \end{equation}
  be the sorted positive finite E4M3 values, and let $\iota(\ell)$ be the index of $\ell$ in this list. From $\ell_{r,i}^{\mathrm{init}}$, we take the $D$ nearest smaller values, the initialization itself, and the $U$ nearest larger values:
  \begin{equation}
      \mathcal{D}
      =
      \{-D,\ldots,-1,\,0,\,+1,\ldots,+U\},
      \qquad
      D = K-1-U,
      \qquad
      |\mathcal{D}| = K,
      \label{eq:shift_set}
  \end{equation}
  which defines the \emph{local-scale} candidate set
  \begin{equation}
      \mathcal{L}_{r,i}
      =
      \Bigl\{\,
      \lambda_{\;\mathrm{clip}\left(\,\iota(\ell_{r,i}^{\mathrm{init}})+\delta,\;1,\;M\,\right)}
      \;:\;
      \delta\in\mathcal{D}
      \,\Bigr\}.
      \label{eq:fp8_local_walk}
  \end{equation}
  Thus $\delta>0$ selects a larger scale, $\delta<0$ a smaller one, and $\delta=0$ recovers the initialization. Throughout the paper we use $U=6$ and $K=16$, i.e.\ nine steps down and six steps up. Shifts that would leave the grid are clipped to the endpoints. Every candidate is exactly representable in E4M3 by construction.

  Mapping each local scale back through the global factor yields the \emph{effective-scale} candidate set used in Eq.~\eqref{eq:scale_objective}:
  \begin{equation}
      \mathcal{S}_{r,i}
      =
      \bigl\{
      s_{\mathrm{g}}\,\ell
      \ \big|\
      \ell\in\mathcal{L}_{r,i}
      \bigr\}.
      \label{eq:fp8_candidate_set}
  \end{equation}
  Each candidate $\tilde{s}\in\mathcal{S}_{r,i}$ is evaluated by standard FP4 group quantization,
  \begin{equation}
      \hat{\mathbf{w}}_{r,i}(\tilde{s}) =
      Q_{\mathrm{fp4}}\!\left(\frac{\mathbf{w}_{r,i}}{\tilde{s}}\right)\tilde{s},
  \end{equation}
  and the best hardware-valid scale is chosen independently for each group:
  \begin{equation}
      \tilde{s}_{r,i}^{*}
      =
      \arg\min_{\tilde{s}\in\mathcal{S}_{r,i}}
      \left\|
      \left(\mathbf{w}_{r,i} - \hat{\mathbf{w}}_{r,i}(\tilde{s})\right)
      \odot \sqrt{\mathbf{h}_{i}}
      \right\|_2^2.
      \label{eq:scale_argmin}
  \end{equation}
  When $\mathbf{h}_{i}=\mathbf{1}$, the same procedure reduces to the scale-only search in our ablations. 

  \paragraph{Final re-quantization.}
  After the best hardware-valid scale $\tilde{s}_{r,i}^{*}$ is chosen, we perform one final FP4 quantization with that scale. The search is offline and fully separable: different groups, and the $K$ candidates inside each group, can be scored independently, so the inner loop batches on GPU as FP4 projection plus element-wise weighted error. Relative to weight-update methods such as GPTQ, the extra cost is small, and inference is unchanged. This is why H-Scale can sit as a drop-in refinement on RTN, 4over6, and other NVFP4 pipelines.

  \begin{figure*}[t]
    \centering
  \begin{minipage}[t]{0.48\textwidth}
  \centering
  \includegraphics[width=\linewidth]{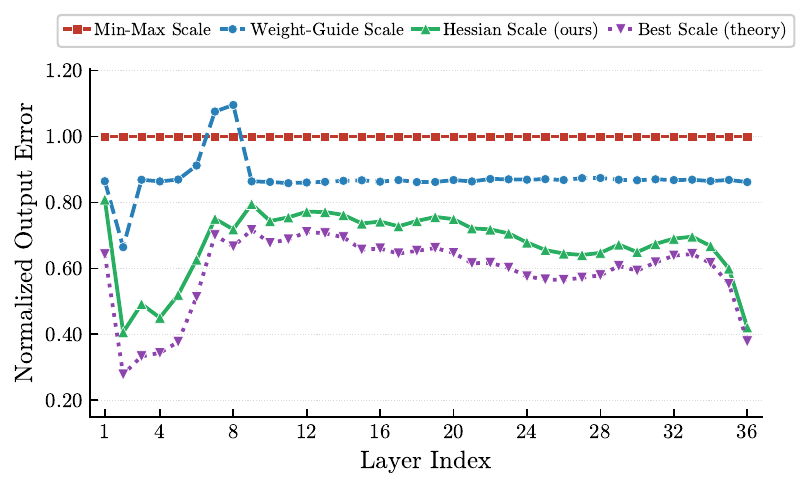}
  \centerline{(a) Normalized layer output error}
  \end{minipage}
    \hfill
  \begin{minipage}[t]{0.48\textwidth}
    \centering
  \includegraphics[width=\linewidth]{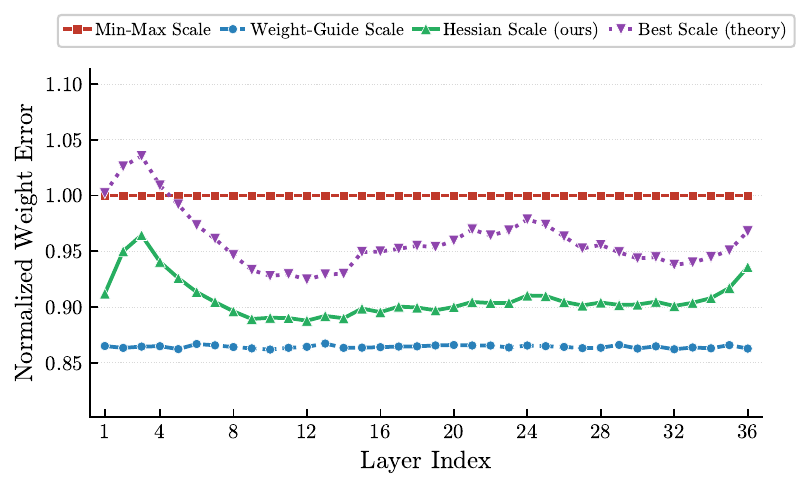}
  \centerline{(b) Normalized weight MSE}
  \end{minipage}
  \caption{\textbf{Gate-projection layer-wise validation on Qwen3-4B-Instruct.} Both metrics are normalized by the Min-Max baseline at each layer. Although Weight-Guided Scale gives the lowest unweighted weight MSE, H-Scale closely tracks the Best Scale oracle for layer output error, demonstrating the mismatch between weight reconstruction and output preservation.}
  \label{fig:gate_layer_validation}
    \end{figure*}
    
\subsection{Why Hessian-Guided Scale Search Works}
\label{sec:why_hscale_works}
  
Traditional NVFP4 RTN treats the group scale as a range-fitting parameter. For each $1{\times}16$ weight group, it chooses a min-max (or max-abs) scale so that all values fall within the representable FP4 range, and then rounds every weight to the induced E2M1 grid. Min-max scaling only fits that range. It ignores both the rounding of the coarse FP4 grid and the distribution of the remaining weights.

A single outlier can therefore determine the scale for all 16 values and decide which FP4 codes are actually used by the majority of weights. Scale-refinement methods such as SignRoundV2~\citep{signroundv2scale} and MR-GPTQ~\citep{egiazarian2025bridginggappromiseperformance} improve on this by searching the initial scale to minimize $\|\mathbf{w}-\hat{\mathbf{w}}\|_2^2$. This is exactly a special case of Eq.~\eqref{eq:scale_objective} with $\mathbf{h}_{i}=\mathbf{1}$, i.e., every coordinate in the group is treated as equally important. H-Scale keeps the same hardware-valid NVFP4 scale space, but replaces this identity weighting with the diagonal Hessian proxy. As a result, the chosen scale allocates FP4 resolution toward weights that matter most for the calibration-layer output.


Figure~\ref{fig:gate_layer_validation} evaluates this distinction on all 36 \texttt{gate\_proj} layers of Qwen3-4B-Instruct, with every curve normalized by the Min-Max baseline at the same layer (red line $\equiv 1.0$). This validation uses 4096 FineWeb~\citep{penedo2024fineweb} calibration samples with sequence length 8192. We report both layer-output error and unweighted weight MSE to test whether the diagonal-Hessian objective in Eq.~\eqref{eq:scale_objective} is a better surrogate for Eq.~\eqref{eq:layer_output_objective} than plain reconstruction error.

\emph{Min-max scaling} is the standard RTN range-fitting rule above. \emph{Weight-centric scale search} (``Weight-Guided Scale'' in the plot) searches the same scale candidates as H-Scale, but minimizes unweighted $\|\mathbf{w}-\hat{\mathbf{w}}\|_2^2$ by setting $\mathbf{h}_{i}=\mathbf{1}$. \emph{Best Scale} is an oracle reference obtained by backpropagating through the scale variables to optimize the full layer-output objective in Eq.~\eqref{eq:layer_output_objective}.

Relative to Min-Max, H-Scale reduces activation loss by 32.5\% on average, close to the 40.8\% reduction of Best Scale (about 80\% of the oracle gain). Best Scale fine-tunes continuous scale variables by gradient descent and is much more expensive than the discrete search used at PTQ time. Weight-Guided Scale attains the lowest unweighted weight MSE, yet reduces activation loss by only 12.6\%. Therefore, H-Scale approaches the oracle activation-loss reduction through a lightweight Hessian-guided search. 


  \subsection{Integration with NVFP4 SOTA: H-Scale as an RTN Replacement}
  
  Most NVFP4 pipelines end with the same local step: pick a per-group scale, typically by max-abs / RTN, and round the group to the E2M1 grid. H-Scale replaces only that pick. Thus, \textbf{any NVFP4 pipeline whose final local projection 
  uses an RTN-style scale selector can replace that selector with H-Scale} without changing the 
  rest of the algorithm. This yields a unified view of how H-Scale plugs into existing SOTA.
  
  Methods such as 4over6~\citep{cook2026six} and ArcQuant~\citep{meng2026arcquant} first optimize structure (e.g., 4/6 mapping, residual channels), then perform per-group quantization. H-Scale leaves those structural decisions intact and replaces the final local scale selection: start from the baseline's chosen scale, run the Hessian-weighted search over the effective-scale candidates in Eq.~\eqref{eq:fp8_candidate_set}, and re-quantize with the selected hardware-valid effective scale.
  
  GPTQ~\citep{frantar2022gptq} and variants (GPTAQ, MR-GPTQ) use the full Hessian $\mathbf{H}$ for column ordering and for propagating rounding error across columns via $\mathbf{H}^{-1}$. The local projection to the NVFP4 grid, however, still requires choosing a group scale and rounding to the grid. H-Scale replaces this local scale-selection step: for each group, it uses $\mathbf{h} = \text{diag}(\mathbf{H})$ in the scale-search objective Eq.~\eqref{eq:scale_objective}. GPTQ keeps its ordering and error propagation, and only the scale selection becomes Hessian-guided. H-Scale is therefore a drop-in replacement for the RTN-style scale-selection step inside these pipelines.
  
  A single calibration pass yields $\mathbf{h}$ per layer, or reuses $\mathrm{diag}(\mathbf{H})$ when GPTQ already built it. The same replacement then applies to RTN, 4over6, ArcQuant, and GPTQ-like methods. The NVFP4 format and inference cost do not change.

  \section{Experiments}
  \label{sec:experiments}
  
  \begin{table*}[t]
  \centering
  \caption{Performance evaluation on Qwen3-30A3-Thinking. H-Scale generally improves average performance across PTQ pipelines, though not every individual benchmark improves for every method.}
  \label{tab:qwen30a3}
  \begin{tabular}{@{}lcccccc@{}}
  \toprule
  \textbf{Method} & \textbf{Avg.} & \textbf{AIME24} & \textbf{AIME25} & \textbf{MMLU-R} & \textbf{LiveBench} & \textbf{LCBench} \\ \midrule
  BF16 Baseline & 81.06 & 89.52 & 78.10 & 90.85 & 76.23 & 70.58 \\ \midrule
  RTN & 80.03 & 86.76 & 77.61 & 90.54 & \textbf{75.14} & 70.12 \\
  \quad + H-Scale & \textbf{80.68} & \textbf{89.33} & \textbf{77.98} & \textbf{90.65} & 74.72 & \textbf{70.74} \\ \midrule
  4over6 & 79.99 & 87.44 & 77.04 & 90.23 & 75.04 & 70.20 \\
  \quad + H-Scale & \textbf{80.49} & \textbf{87.82} & \textbf{77.84} & \textbf{90.72} & \textbf{75.53} & \textbf{70.56} \\ \midrule
  ArcQuant & 79.73 & 87.91 & 74.53 & 90.42 & \textbf{75.74} & \textbf{70.04} \\
  \quad + H-Scale & \textbf{80.46} & \textbf{89.03} & \textbf{77.78} & \textbf{90.89} & 74.65 & 69.97 \\ \midrule
  GPTQ & 80.01 & 86.67 & 78.76 & \textbf{90.77} & 74.16 & 69.71 \\
  \quad + H-Scale & \textbf{81.22} & \textbf{88.94} & \textbf{80.45} & 90.07 & \textbf{76.13} & \textbf{70.49} \\ \midrule
  GPTAQ & 80.21 & 87.63 & 78.32 & \textbf{90.79} & 74.33 & \textbf{69.99} \\
  \quad + H-Scale & \textbf{80.71} & \textbf{88.42} & \textbf{78.65} & 90.71 & \textbf{76.09} & 69.69 \\ \midrule
  MR-GPTQ & 79.67 & 85.35 & \textbf{78.00} & \textbf{90.81} & 75.25 & 68.93 \\
  \quad + H-Scale & \textbf{80.41} & \textbf{87.00} & 77.95 & 90.59 & \textbf{76.03} & \textbf{70.49} \\ \bottomrule
  \end{tabular}
  \end{table*}
  
  \begin{table*}[t]
  \centering
  \caption{Evaluation on Qwen3-30A3-Instruct. H-Scale usually improves average accuracy across eight diverse tasks, with mixed effects on individual benchmarks.}
  \label{tab:qwen30a3_instruct}
  \resizebox{0.98\textwidth}{!}{%
  \begin{tabular}{@{}lccccccccc@{}}
  \toprule
  \textbf{Method} & \textbf{Avg.} & \textbf{C-Eval} & \textbf{LiveBench} & \textbf{MMLU-R} & \textbf{AIME25} & \textbf{LCBench} & \textbf{ARC-C} & \textbf{BBH} & \textbf{GPQA} \\ \midrule
  BF16 Baseline & 65.58 & 63.43 & 69.27 & 89.11 & 59.90 & 48.42 & 69.05 & 70.50 & 54.99 \\ \midrule
  RTN & 64.04 & 62.51 & \textbf{68.34} & \textbf{89.23} & 56.91 & 47.56 & 67.35 & \textbf{68.33} & 52.07 \\
  \quad + H-Scale & \textbf{64.78} & \textbf{62.73} & 68.23 & 89.14 & \textbf{57.69} & \textbf{47.61} & \textbf{68.10} & 67.47 & \textbf{57.24} \\ \midrule
  4over6 & 64.25 & 62.44 & 68.04 & 88.47 & \textbf{57.88} & 47.63 & \textbf{68.50} & 67.96 & 53.11 \\
  \quad + H-Scale & \textbf{64.65} & \textbf{62.77} & \textbf{68.49} & \textbf{89.19} & 57.65 & \textbf{48.19} & 68.37 & \textbf{68.03} & \textbf{54.54} \\ \midrule
  ArcQuant & 63.45 & 62.59 & 66.82 & \textbf{89.02} & 55.03 & 46.55 & \textbf{68.16} & 67.45 & 52.00 \\
  \quad + H-Scale & \textbf{64.24} & \textbf{63.03} & \textbf{67.24} & 88.70 & \textbf{55.68} & \textbf{47.13} & 67.58 & \textbf{67.49} & \textbf{57.04} \\ \midrule
  GPTQ & 63.90 & \textbf{63.29} & 68.84 & \textbf{88.74} & 56.22 & \textbf{48.43} & 67.31 & 66.98 & 51.38 \\
  \quad + H-Scale & \textbf{64.50} & 62.45 & \textbf{69.29} & 88.59 & \textbf{56.37} & 47.44 & \textbf{68.40} & \textbf{68.38} & \textbf{55.11} \\ \midrule
  GPTAQ & 64.27 & \textbf{62.65} & \textbf{69.12} & \textbf{89.22} & \textbf{58.55} & 47.22 & 67.61 & \textbf{67.70} & 52.11 \\
  \quad + H-Scale & \textbf{64.74} & 62.62 & 69.10 & 88.82 & 58.02 & \textbf{48.68} & \textbf{68.68} & 67.38 & \textbf{54.61} \\ \midrule
  MR-GPTQ & 63.63 & 62.28 & \textbf{68.02} & 88.53 & 55.68 & 47.68 & \textbf{68.39} & \textbf{68.15} & 50.34 \\
  \quad + H-Scale & \textbf{64.48} & \textbf{62.66} & 67.41 & \textbf{88.75} & \textbf{59.85} & \textbf{47.82} & 68.07 & 67.29 & \textbf{54.00} \\ \bottomrule
  \end{tabular}%
  }
  \end{table*}
  
  \begin{table*}[t]
  \centering
  \caption{Evaluation on Qwen3-4B-Instruct. H-Scale yields higher observed average scores across the tested NVFP4 pipelines, while individual benchmark gains remain method-dependent.}
  \label{tab:qwen3-4b}
  \resizebox{0.98\textwidth}{!}{%
  \begin{tabular}{@{}lccccccccc@{}}
  \toprule
  \textbf{Method} & \textbf{Avg.} & \textbf{C-Eval} & \textbf{LiveBench} & \textbf{MMLU-R} & \textbf{AIME25} & \textbf{LCBench} & \textbf{ARC-C} & \textbf{BBH} & \textbf{GPQA} \\ \midrule
  BF16 Baseline & 57.76 & 55.54 & 63.16 & 83.98 & 46.16 & 36.63 & 60.72 & 73.50 & 42.41 \\ \midrule
  RTN & 55.38 & \textbf{52.48} & 60.66 & 82.87 & 40.37 & 34.32 & 60.28 & 71.76 & 40.29 \\
  \quad + H-Scale & \textbf{56.26} & 52.43 & \textbf{61.08} & \textbf{83.13} & \textbf{43.24} & \textbf{35.65} & \textbf{60.69} & \textbf{72.39} & \textbf{41.44} \\ \midrule
  4over6 & 55.38 & 52.20 & \textbf{60.43} & \textbf{82.74} & 43.35 & \textbf{36.16} & \textbf{60.10} & 70.16 & 37.91 \\
  \quad + H-Scale & \textbf{56.12} & \textbf{54.70} & 59.95 & 82.73 & \textbf{43.51} & 35.46 & 59.30 & \textbf{72.40} & \textbf{40.95} \\ \midrule
  ArcQuant & 54.49 & 52.42 & 58.87 & 81.27 & 40.19 & \textbf{33.80} & 58.41 & 70.58 & \textbf{40.42} \\
  \quad + H-Scale & \textbf{55.00} & \textbf{53.46} & \textbf{59.63} & \textbf{81.39} & \textbf{42.97} & 33.06 & \textbf{58.56} & \textbf{71.26} & 39.63 \\ \midrule
  GPTQ & 55.63 & 52.42 & \textbf{61.86} & 82.51 & \textbf{42.18} & \textbf{35.24} & 59.87 & 68.17 & 42.78 \\
  \quad + H-Scale & \textbf{56.53} & \textbf{55.11} & 60.39 & \textbf{82.87} & 42.07 & 34.70 & \textbf{60.29} & \textbf{71.23} & \textbf{45.61} \\ \midrule
  GPTAQ & 56.63 & 53.13 & \textbf{62.19} & \textbf{83.44} & \textbf{45.72} & \textbf{36.47} & 60.06 & 71.00 & 41.02 \\
  \quad + H-Scale & \textbf{57.52} & \textbf{54.17} & 61.68 & 83.07 & 43.77 & 36.17 & \textbf{60.68} & \textbf{73.05} & \textbf{47.54} \\ \midrule
  MR-GPTQ & 56.41 & 53.54 & 62.57 & \textbf{83.16} & 44.28 & \textbf{35.13} & 60.33 & 70.89 & 41.38 \\
  \quad + H-Scale & \textbf{56.74} & \textbf{54.86} & \textbf{62.84} & 82.38 & \textbf{44.50} & 34.52 & \textbf{60.84} & \textbf{71.30} & \textbf{42.71} \\ \bottomrule
  \end{tabular}%
  }
  \end{table*}
  
  \begin{table*}[t]
  \centering
  \caption{Cross-family evaluation on LLaMA-3.1-8B-Instruct. Avg. is computed over the seven listed benchmarks. H-Scale yields higher observed average scores for all tested NVFP4 baselines.}
  \label{tab:llama31}
  \resizebox{0.98\textwidth}{!}{%
  \begin{tabular}{@{}lcccccccc@{}}
  \toprule
  \textbf{Method} & \textbf{Avg.} & \textbf{C-Eval} & \textbf{LiveBench} & \textbf{MMLU-R} & \textbf{ARC-C} & \textbf{GSM8K} & \textbf{BBH} & \textbf{GPQA} \\ \midrule
  BF16 Baseline & 53.02 & 51.05 & 19.24 & 66.91 & 51.83 & 76.89 & 68.24 & 36.97 \\ \midrule
  RTN & 48.79 & 45.64 & 18.17 & 63.07 & 49.08 & 70.92 & \textbf{62.41} & \textbf{32.21} \\
  \quad + H-Scale & \textbf{49.53} & \textbf{48.15} & \textbf{18.77} & \textbf{64.30} & \textbf{49.70} & \textbf{72.52} & 61.43 & 31.84 \\ \midrule
  4over6 & 48.86 & 46.70 & 18.63 & 61.68 & 49.73 & 72.19 & 62.24 & 30.87 \\
  \quad + H-Scale & \textbf{50.05} & \textbf{48.21} & \textbf{19.16} & \textbf{63.16} & \textbf{50.21} & \textbf{74.28} & \textbf{64.02} & \textbf{31.31} \\ \midrule
  ArcQuant & 48.40 & 46.30 & 17.92 & \textbf{62.74} & 49.94 & \textbf{72.84} & 60.32 & \textbf{28.72} \\
  \quad + H-Scale & \textbf{48.64} & \textbf{49.19} & \textbf{18.23} & 62.01 & \textbf{51.06} & 71.25 & \textbf{61.09} & 27.64 \\ \midrule
  GPTQ & 50.61 & 48.93 & 19.20 & \textbf{65.32} & 51.73 & 70.91 & 63.57 & 34.63 \\
  \quad + H-Scale & \textbf{51.32} & \textbf{50.28} & \textbf{19.45} & 63.00 & \textbf{51.95} & \textbf{75.33} & \textbf{64.08} & \textbf{35.17} \\ \midrule
  GPTAQ & 49.54 & 46.02 & 17.82 & 64.49 & 50.20 & 70.29 & \textbf{64.22} & 33.74 \\
  \quad + H-Scale & \textbf{51.48} & \textbf{50.61} & \textbf{19.73} & \textbf{65.05} & \textbf{50.84} & \textbf{73.74} & 63.75 & \textbf{36.65} \\ \midrule
  MR-GPTQ & 49.10 & 47.94 & 16.93 & 64.23 & \textbf{51.78} & 67.31 & 63.88 & 31.60 \\
  \quad + H-Scale & \textbf{50.11} & \textbf{48.60} & \textbf{18.60} & \textbf{64.31} & 50.20 & \textbf{68.76} & \textbf{64.28} & \textbf{36.04} \\ \bottomrule
  \end{tabular}%
  }
  \end{table*}

  \begin{table*}[t]
    \centering
    \caption{Ablation on Qwen3-30A3-Instruct: scale with vs.\ without Hessian. ``Scale-only'' uses identity weighting in Eq.~\eqref{eq:scale_objective}.}
    \label{tab:ablation_hessian_instruct}
    \resizebox{0.98\textwidth}{!}{%
    \begin{tabular}{@{}llccccccccc@{}}
    \toprule
    \textbf{Baseline} & \textbf{Variant} & \textbf{Avg.} & \textbf{C-Eval} & \textbf{LiveBench} & \textbf{MMLU-R} & \textbf{AIME25} & \textbf{LCBench} & \textbf{ARC-C} & \textbf{BBH} & \textbf{GPQA} \\ \midrule
    4over6 & Baseline & 64.25 & 62.44 & 68.04 & 88.47 & \textbf{57.88} & 47.63 & \textbf{68.50} & 67.96 & 53.11 \\
    \quad & Scale-only (w/o $\mathbf{H}$) & 63.55 & \textbf{64.92} & 66.65 & 87.33 & 56.58 & 46.72 & 67.15 & 66.53 & 52.52 \\
    \quad & + H-Scale & \textbf{64.53} & 62.77 & \textbf{68.49} & \textbf{89.19} & 57.65 & \textbf{48.19} & 68.37 & \textbf{68.03} & \textbf{53.54} \\ \midrule
    ArcQuant & Baseline & 63.45 & 62.59 & 66.82 & \textbf{89.02} & 55.03 & 46.55 & \textbf{68.16} & 67.45 & 52.00 \\
    \quad & Scale-only (w/o $\mathbf{H}$) & 63.10 & 62.98 & 66.55 & 87.54 & 54.58 & 47.06 & 66.81 & 66.80 & 52.51 \\
    \quad & + H-Scale & \textbf{64.24} & \textbf{63.03} & \textbf{67.24} & 88.70 & \textbf{55.68} & \textbf{47.13} & 67.58 & \textbf{67.49} & \textbf{57.04} \\ \midrule
    GPTQ & Baseline & 63.90 & \textbf{63.29} & 68.84 & 88.74 & 56.22 & \textbf{48.43} & 67.31 & 66.98 & 51.38 \\
    \quad & Scale-only (w/o $\mathbf{H}$) & 64.39 & 62.76 & 67.51 & \textbf{89.24} & \textbf{56.67} & 47.92 & \textbf{68.43} & 67.52 & 55.05 \\
    \quad & + H-Scale & \textbf{64.50} & 62.45 & \textbf{69.29} & 88.59 & 56.37 & 47.44 & 68.40 & \textbf{68.38} & \textbf{55.11} \\ \bottomrule
    \end{tabular}}
    \end{table*}

        Our experiments ask whether H-Scale raises the average performance of strong NVFP4 baselines, transfers across model scales and families, and provides enough benefit to justify its one-time offline Hessian cost.
  
        \subsection{Experimental Setup}
        \noindent\textbf{Benchmarks.} We use AIME, MMLU-Redux, LiveBench, LiveCodeBench, C-Eval, ARC-Challenge (ARC-C), BBH, GPQA, and GSM8K to evaluate reasoning, math, knowledge, and coding ability~\citep{clark2018think,cobbe2021gsm8k,gema2025mmlu,huang2023ceval,jain2025livecodebench,maa2024aime,rein2023gpqa,suzgun2022bbh,white2024livebench}. Model-specific benchmark suites are detailed in Appendix~\ref{app:evaluation_protocol}.
        
        \noindent\textbf{Decoding.} All methods use a fixed decoding protocol for paired comparisons. Because generation is stochastic, we evaluate each setting with \textbf{three} independent decoding runs and report the mean score. Complete generation settings are provided in Appendix~\ref{app:evaluation_protocol}.
        

        \noindent\textbf{Baselines.} H-Scale is applied to NVFP4 weight-only PTQ pipelines at the hardware-native group size $g{=}16$: RTN, GPTQ~\citep{frantar2022gptq}, 4over6~\citep{cook2026six}, ArcQuant~\citep{meng2026arcquant}, MR-GPTQ~\citep{egiazarian2025bridginggappromiseperformance}, and GPTAQ~\citep{li2025gptaq}. All methods use 4096 FineWeb~\citep{penedo2024fineweb} calibration samples with sequence length 8192. We do not include activation-migration methods such as SmoothQuant~\citep{xiao2023smoothquant} or AWQ~\citep{lin2023awq}, or Hadamard-based preprocessing: $g{=}16$ is fixed by Blackwell tensor-core support, and prior NVFP4 work finds that per-group scales already handle local outliers sufficiently well that these redistribution techniques yield little additional benefit~\citep{egiazarian2025bridginggappromiseperformance}. 
        
        \subsection{Main Results}

  \noindent\textbf{Large Reasoning Model.} Table~\ref{tab:qwen30a3} shows that H-Scale yields a higher average score for every PTQ pipeline on Qwen3-30A3-Thinking. H-Scale also narrows the gap to the BF16 baseline (81.06) for every pipeline, with GPTQ + H-Scale reaching \textbf{81.22}. Observed gains are most consistent on AIME24 and AIME25, suggesting that output-aware scale alignment helps preserve reasoning behavior in NVFP4.
  
  \noindent\textbf{Instruction-Tuned Model.} Table~\ref{tab:qwen30a3_instruct} reports the eight-task suite for Qwen3-30A3-Instruct. H-Scale again yields higher averages for every baseline.
  
  \noindent\textbf{Smaller Model Scaling.} Table~\ref{tab:qwen3-4b} extends the evaluation to Qwen3-4B-Instruct. Improvements are visible on GPQA, BBH, and AIME25, indicating that Hessian-guided scale refinement remains useful at lower model capacity.
  
  \noindent\textbf{Cross-Family Transfer.} Table~\ref{tab:llama31} evaluates LLaMA-3.1-8B-Instruct to test whether the gains are Qwen-specific. H-Scale yields higher average scores for every tested baseline, with especially consistent gains on C-Eval and LiveBench.
  
  \subsection{Ablation Analysis}
  \label{sec:ablation}
  
  \textbf{Is the Hessian necessary?} We compare the raw baseline, a \textit{scale-only} variant that replaces the Hessian diagonal in Eq.~\eqref{eq:scale_objective} with the identity, and full \textit{H-Scale}. Table~\ref{tab:ablation_hessian_instruct} reports the Qwen3-30A3-Instruct ablation. H-Scale gives the highest average score for all three baselines, while scale-only optimization is less stable and can hurt 4over6 and ArcQuant.

  \textbf{Efficiency and Hardware Deployment.} H-Scale preserves the NVFP4 model structure, so inference cost is unchanged. All runtimes in Table~\ref{tab:runtime} are measured on a single NVIDIA H200 GPU. For baselines such as GPTQ that already compute a Hessian, H-Scale only adds the scale search. In Table~\ref{tab:runtime}, GPTQ increases from 9.6\,h to 9.9\,h. Baselines that do not already compute Hessian statistics require a one-time activation pass. The representative 4over6 case increases from 0.2\,h to 0.4\,h, and the offline cost remains modest relative to inference deployment.

\begin{table}[t]
\centering
\caption{Runtime comparison between each baseline and its H-Scale variant using the same hardware and calibration data.}
\label{tab:runtime}
\begin{tabular}{@{}lc@{\hspace{2em}}lc@{}}
\toprule
\textbf{Method} & \textbf{Runtime} & \textbf{Method} & \textbf{Runtime} \\ \midrule
4over6 & 0.2\,h & GPTQ & 9.6\,h \\
4over6 + H-Scale & 0.4\,h & GPTQ + H-Scale & 9.9\,h \\ \bottomrule
\end{tabular}
\end{table}

  \textbf{Selected Scale Shifts.}
  \label{app:scale_shifts}
  We report which shift $\delta$ is selected, and how much Hessian-weighted rounding error each choice removes. All numbers are on Qwen3-4B-Instruct, pooling the three MLP projections ($5.60\times10^{7}$ groups per type; $1.68\times10^{8}$ groups in total). Positive $\delta$ is a larger scale, negative $\delta$ a smaller one, and $\delta=0$ recovers the initialization.

\begin{figure*}[t]
\centering
\begin{minipage}[t]{0.48\textwidth}
\centering
\includegraphics[width=\linewidth]{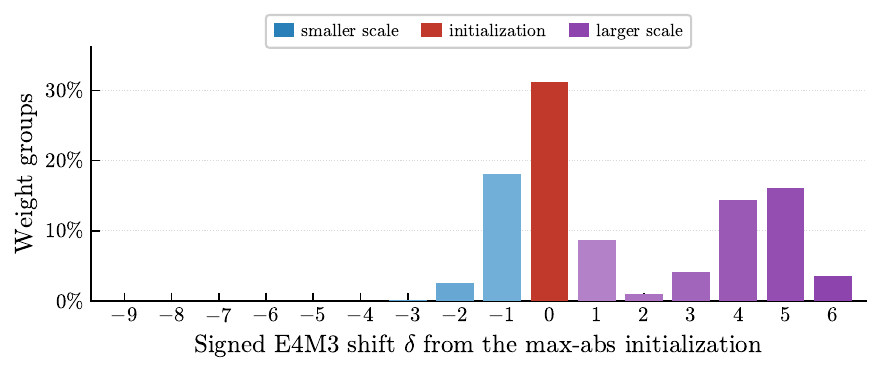}
\centerline{(a) Selected shift $\delta$}
\end{minipage}
\hfill
\begin{minipage}[t]{0.48\textwidth}
\centering
\includegraphics[width=\linewidth]{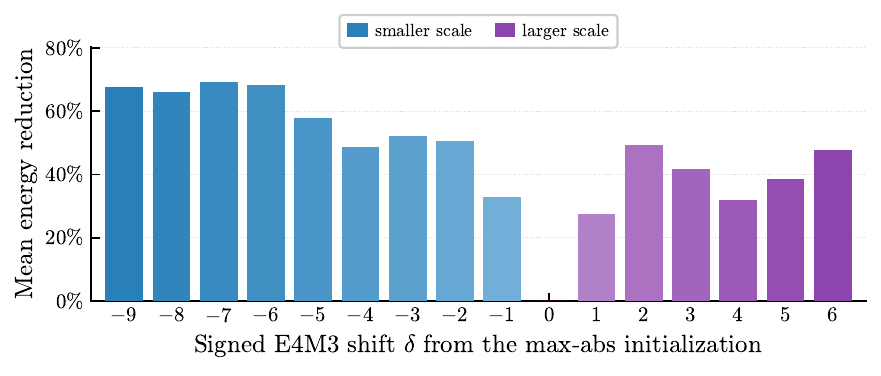}
\centerline{(b) Mean Hessian-weighted error reduction}
\end{minipage}
\caption{\textbf{Scale-shift statistics on the three MLP projections of Qwen3-4B-Instruct} ($1.68\times10^{8}$ groups). Positive $\delta$ is a larger scale. (a)~About half of all groups prefer a scale larger than max-abs. (b)~Mean per-group reduction of the Hessian-weighted rounding error $\sum_j h_j d_j^2$ for groups that selected each $\delta$, relative to $\delta=0$. Nearby shifts give a modest reduction, while rare deep contractions give the largest.}
\label{fig:step_dist}
\end{figure*}

  Figure~\ref{fig:step_dist}(a) shows the selected $\delta$. $31.2\%$ of groups keep the max-abs initialization, $47.9\%$ take a larger scale, and $20.9\%$ a smaller one.

  How often a shift is chosen is not the same as how much error it removes. Figure~\ref{fig:step_dist}(b) reports, for groups that selected each $\delta$, the mean reduction of the Hessian-weighted rounding error $\sum_j h_j d_j^2$ from Eq.~\eqref{eq:scale_objective}, relative to $\delta=0$. Nearby shifts give a modest reduction ($32.8\%$ at $\delta=-1$). Deep contractions ($\delta\leq-4$) cut this error by $58.7\%$ on average. These $0.11\%$ of groups account for $8.3\%$ of the total error H-Scale removes. Over the full window, H-Scale removes $50.6\%$ of the initialization error.

  \textbf{Candidate-Window Width.}
  \label{app:window_width}
  The candidate window of Eq.~\eqref{eq:fp8_candidate_set} is parameterized by an upward budget $U$ and a total budget $K$, giving $D=K-1-U$ downward neighbours. We use $U=6$ and $K=16$ throughout. Table~\ref{tab:window_sweep} sweeps $(U,K)$ over all $144$ linear layers of Qwen3-4B-Instruct. Candidates are still chosen by Eq.~\eqref{eq:scale_objective}. We then evaluate the chosen scales with the true layer-output error $L_{\mathrm{full}}=\mathrm{Tr}(\Delta\mathbf{W}\mathbf{G}\Delta\mathbf{W}^{\top})$, and report how much of the max-abs$\to$full-grid $L_{\mathrm{full}}$ gain each window recovers. A score of $100\%$ means the window matches enumerating every positive E4M3 scale. We summarize the $144$ layers by the median and the $10$th percentile (P10).

  Figure~\ref{fig:step_dist}(a) already indicates that $U$ must cover the $\delta\in\{+4,+5,+6\}$ cluster. For $U\leq 5$ it does not: contraction-only ($U=0$) recovers only $64.63\%$ at the median. For $U\geq 6$ the median saturates at $100.00\%$ and P10 remains above $99.6\%$, so a $K{=}16$ window already matches the full-grid improvement on essentially every layer. We therefore take the smallest such $U$, leaving $D=9$ downward steps. Widening to $K=20$ or $24$ changes the recovered fraction only marginally, so extra downward budget yields little additional gain.

\begin{table}[t]
  \centering
  \caption{Candidate-window sweep on all $144$ linear layers of Qwen3-4B-Instruct. \textbf{Median} and \textbf{P10} are the layer-wise fraction (\%) of the max-abs$\to$full-E4M3 $L_{\mathrm{full}}$ gain recovered by each $(U,K)$ window. Default in bold. $\dagger$ marks a wider budget.}
  \label{tab:window_sweep}
  \begin{tabular*}{0.70\textwidth}{@{\extracolsep{\fill}}rccc@{}}
  \toprule
  \textbf{$U$ / $K$} & \textbf{$\delta$ range} & \textbf{Median (\%)} & \textbf{P10 (\%)} \\
  \midrule
  \phantom{0}0 / 16 & $[-15,\phantom{+0}0]$ & \phantom{0}64.63 & 57.94 \\
  \phantom{0}4 / 16 & $[-11,\phantom{0}+4]$ & \phantom{0}90.29 & 87.46 \\
  \phantom{0}5 / 16 & $[-10,\phantom{0}+5]$ & \phantom{0}98.59 & 97.83 \\
  \textbf{\phantom{0}6 / 16} & $\mathbf{[\phantom{0}-9,\phantom{0}+6]}$ & \textbf{100.00} & \textbf{99.91} \\
  \phantom{0}7 / 16 & $[\phantom{0}-8,\phantom{0}+7]$ & 100.00 & 99.89 \\
  \phantom{0}8 / 16 & $[\phantom{0}-7,\phantom{0}+8]$ & 100.00 & 99.84 \\
  10 / 16 & $[\phantom{0}-5,+10]$ & 100.00 & 99.63 \\
  \midrule
  $\dagger$\phantom{0}6 / 20 & $[-13,\phantom{0}+6]$ & 100.00 & 99.96 \\
  $\dagger$\phantom{0}6 / 24 & $[-17,\phantom{0}+6]$ & 100.00 & 99.96 \\
  \bottomrule
  \end{tabular*}
  \end{table}

  \section{Conclusion, Limitations, and Future Work}
  \label{sec:conclusion}
  
  We identify per-group scale selection as an important bottleneck for NVFP4 fidelity and propose \textbf{H-Scale}, a diagonal-Hessian-guided method that chooses group scales by a weighted reconstruction of the layer output. As a plug-and-play post-processing step, H-Scale generally yields higher average performance across diverse NVFP4 baselines, model sizes, and model families, including both Qwen and LLaMA evaluations.
  
  Our study is limited to NVFP4 with $g=16$ and text-only LLMs. Whether the same scale-contraction strategy transfers to other formats, group sizes, or multimodal models is left open. Along these lines, we plan to explore scale tuning for multimodal architectures.

  {\small
  \bibliographystyle{plainnat}
  \bibliography{refs}
  }
  
\appendix
\raggedbottom
  
\newpage

\section{Additional Analysis for Hessian-Guided Scale Search}
\label{app:hscale_analysis}



\subsection{Calibration Microcase}
\label{app:scale_microcase}

We first use a small calibration example to isolate why an output-aware scale objective can prefer a different FP4 code pattern from weight-centric search. Table~\ref{tab:scale_search_microcase} makes the objective mismatch explicit in a 4-dimensional calibration microcase with
$\mathbf{w}=[-3.750,-1.740,4.320,4.880]$ and two calibration activations
\[
\mathbf{X}=
\begin{bmatrix}
0.600 & 1.400 & 1.500 & 2.000\\
2.000 & 1.000 & 0.700 & 1.000
\end{bmatrix}.
\]
The output error is $\|\mathbf{X}(\mathbf{w}-\hat{\mathbf{w}})^{\top}\|_2^2$.

\begin{table}[H]
\centering
\small
\caption{A calibration microcase showing why scale search should be output-aware.}
\label{tab:scale_search_microcase}
\begin{tabular*}{\textwidth}{@{\extracolsep{\fill}}lcccc@{}}
\toprule
\textbf{Scale choice} & \textbf{Objective} & \textbf{Scale} & $\hat{\mathbf{w}}$ & \textbf{Weight / Output error} \\
\midrule
Min-max & Preserve range & 0.813 & $[-3.253,-1.627,4.880,4.880]$ & $0.573$ / $3.927$ \\
Weight-centric search & $ \min \|\mathbf{w}-\hat{\mathbf{w}}\|_2^2$ & 0.801 & $[-3.204,-1.602,4.805,4.805]$ & $\mathbf{0.559}$ / $3.448$ \\
Hessian-guided search &  $ \min \|\mathbf{X}(\mathbf{w}-\hat{\mathbf{w}})^{\top}\|_2^2$ & 0.895 & $[-3.579,-1.789,3.579,5.368]$ & $0.820$ / $\mathbf{0.079}$ \\
\bottomrule
\end{tabular*}
\end{table}

Min-max scaling is set by the largest-magnitude weight, so that coordinate fixes the grid for the whole group. Weight-centric search then finds the scale with the smallest unweighted reconstruction error, treating all four coordinates equally. Scoring candidates by layer-output error selects a different FP4 code pattern, $[-4,-2,4,6]$, and a larger unweighted error ($0.820$ vs.\ $0.559$). The output error falls from $3.927$ (min-max) and $3.448$ (weight-centric) to $0.079$, because the residual is moved off the high-impact activation directions. H-Scale approximates this output-aware score with the diagonal Hessian proxy. The layer-wise plots below repeat the comparison at model scale.

\subsection{Additional Layer-Wise Validation}
\label{app:additional_layer_validation}

\begin{figure}[H]
\centering
\begin{minipage}[t]{0.48\textwidth}
\centering
\includegraphics[width=\linewidth]{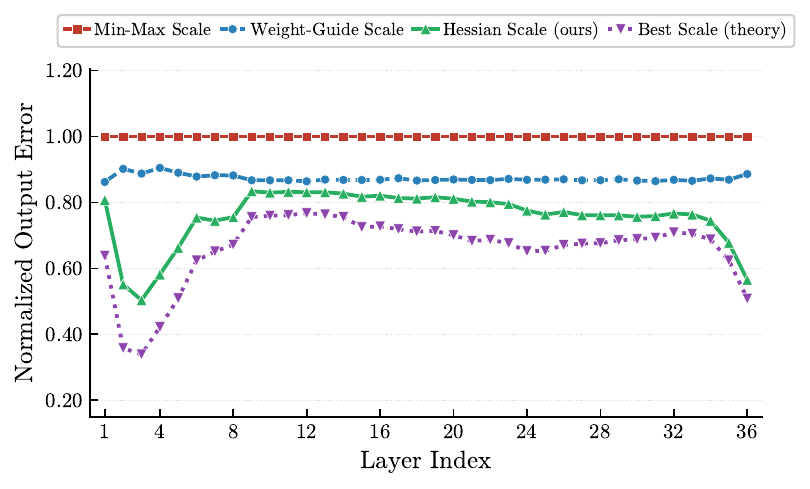}
\centerline{(a) \texttt{up\_proj} output error}
\end{minipage}
\hfill
\begin{minipage}[t]{0.48\textwidth}
\centering
\includegraphics[width=\linewidth]{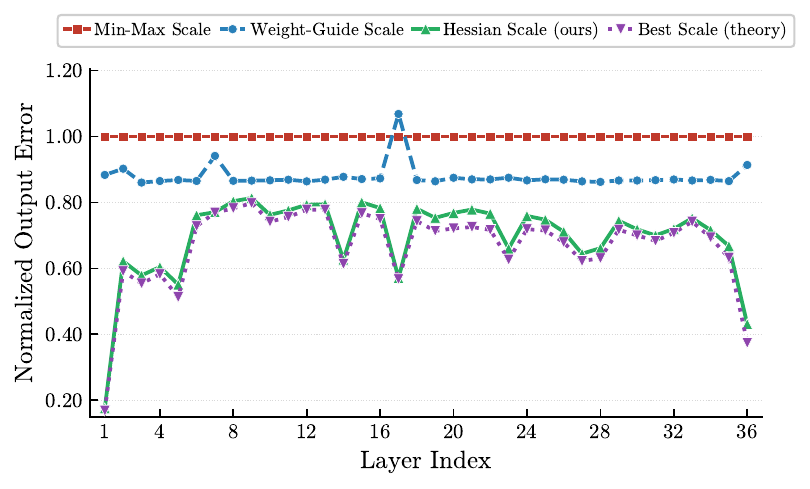}
\centerline{(b) \texttt{down\_proj} output error}
\end{minipage}
\caption{\textbf{Additional normalized layer output error on Qwen3-4B-Instruct.} All curves are normalized by the Min-Max baseline at each layer. Best Scale is the oracle reference and gives the lowest output error, while H-Scale is the strongest practical scale-search method, reducing average error over Min-Max by 24\% for \texttt{up\_proj} and 30\% for \texttt{down\_proj}.}
\label{fig:additional_oe_comparison}
\end{figure}

\begin{figure}[H]
\centering
\begin{minipage}[t]{0.48\textwidth}
\centering
\includegraphics[width=\linewidth]{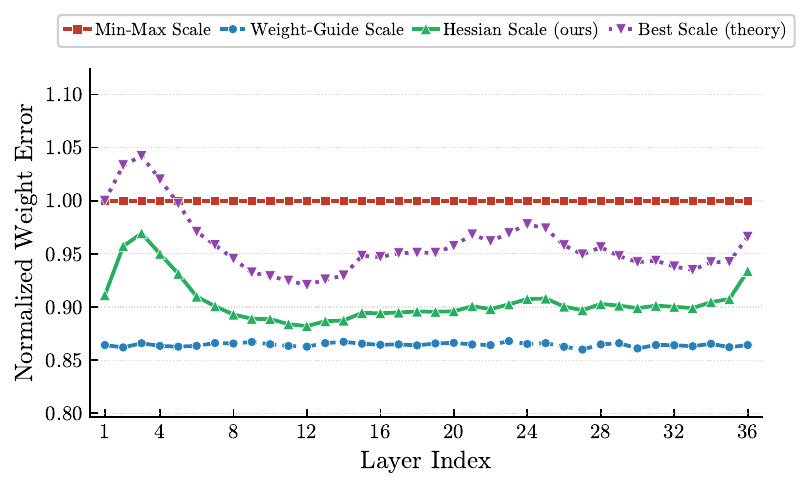}
\centerline{(a) \texttt{up\_proj} weight MSE}
\end{minipage}
\hfill
\begin{minipage}[t]{0.48\textwidth}
\centering
\includegraphics[width=\linewidth]{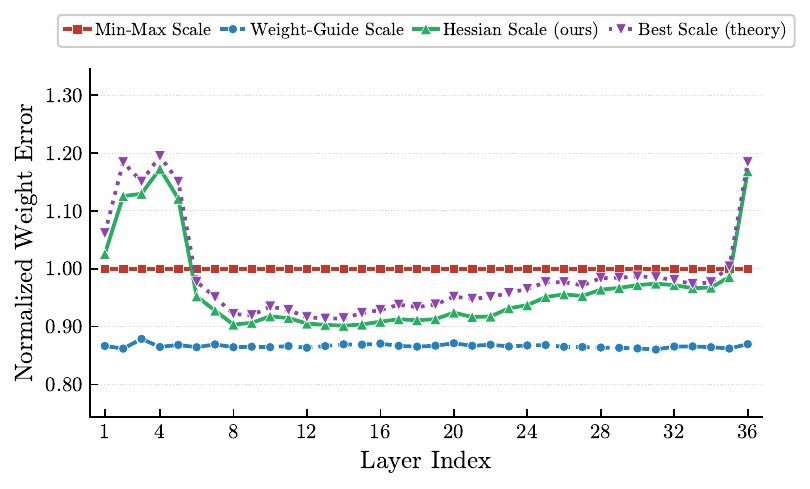}
\centerline{(b) \texttt{down\_proj} weight MSE}
\end{minipage}
\caption{\textbf{Additional normalized weight MSE on Qwen3-4B-Instruct.} Weight-Guided Scale generally minimizes unweighted reconstruction error, while H-Scale gives lower output error. For \texttt{down\_proj}, H-Scale can exceed the Min-Max weight MSE at early layers and the final layer while still reducing output error, confirming that weight MSE is an unreliable proxy for layer output fidelity.}
\label{fig:additional_mse_comparison}
\end{figure}

The microcase above isolates the mechanism at the group level. We next check whether the same output-aware behavior appears consistently across layers. The layer-wise trend shown for \texttt{gate\_proj} in Figure~\ref{fig:gate_layer_validation} also holds for the other MLP projections, so we report the additional \texttt{up\_proj} and \texttt{down\_proj} curves in Figures~\ref{fig:additional_oe_comparison} and~\ref{fig:additional_mse_comparison}.

\section{Implementation Pseudocode}
\label{app:pseudocode}

After the empirical validation, we spell out the implementation used in the experiments. Algorithms~\ref{alg:hscale}--\ref{alg:gptq_hscale} summarize the standalone scale search and its integration into representative NVFP4 pipelines. The diagonal weights $\mathbf{h}$ are computed as $\mathrm{diag}(\mathbf{X}^{\top}\mathbf{X})$ from calibration activations, or reused from the Hessian statistics already available in GPTQ-like baselines. H-Scale only refines the per-group scale and keeps the NVFP4 grouping, FP4 value grid, and FP8 scale representation unchanged. In Algorithms~\ref{alg:four_over_six_hscale} and~\ref{alg:gptq_hscale}, the red lines mark the modules newly added on top of the original baseline procedure.

The search is fully parallelizable. Different row-group pairs and different candidates in $\mathcal{S}_{r,i}$ are independent. The pseudocode below makes that data-parallel structure explicit.

For GPTQ, H-Scale is inserted only at the local quantization step: the vanilla call $\mathbf{q}=Q(\mathbf{w})$ is replaced by a Hessian-weighted NVFP4 scale search, while the blockwise GPTQ error computation and propagation remain unchanged.

\begin{algorithm}[H]
\caption{ \textbf{HScaleGroupQuantize}: H-Scale for NVFP4 scale refinement}
\label{alg:hscale}
\begin{algorithmic}[1]
\Require Weight matrix $\mathbf{W}$, diagonal Hessian weights $\mathbf{h}$, initial effective scales $\mathbf{s}^{\mathrm{init}}$, group size $g=16$, upward budget $U=6$, candidate budget $K=16$, global scale $s_{\mathrm{g}}$, E4M3 code count $M$
\Ensure Quantized weight matrix $\hat{\mathbf{W}}$
\ForAll{row-group pairs $(r,i)$}
    \State $\mathbf{w}\gets \mathbf{w}_{r,i}$, $\mathbf{h}_i\gets$ aligned slice of $\mathbf{h}$
    \State $s^{\mathrm{init}}\gets$ aligned baseline-provided effective NVFP4 scale for $\mathbf{w}$
    \State $\ell^{\mathrm{init}}\gets Q_{\mathrm{fp8}}(s^{\mathrm{init}}/s_{\mathrm{g}})$, \quad $c\gets \iota(\ell^{\mathrm{init}})$ \Comment{$\iota$: E4M3 code index, i.e.\ the stored scale byte}
    \State $D\gets K-1-U$ \Comment{downward budget}
    \For{$\delta=-D,\ldots,U$}
        \State $\ell^{(\delta)}\gets \lambda_{\mathrm{clip}(c+\delta,\,1,\,M)}$ \Comment{$\delta>0$: larger scale; $\delta=0$: $\ell^{\mathrm{init}}$; $\delta<0$: smaller}
    \EndFor
    \State $\mathcal{L}\gets\bigl\{\ell^{(\delta)} \bigm| -D\le\delta\le U\bigr\}$
    \State $\mathcal{S}\gets \bigl\{ s_{\mathrm{g}}\,\ell \bigm| \ell\in\mathcal{L} \bigr\}$
    \ForAll{$\tilde{s}\in\mathcal{S}$}
        \State $\hat{\mathbf{w}}(\tilde{s})\gets Q_{\mathrm{fp4}}(\mathbf{w}/\tilde{s})\,\tilde{s}$
        \State $E(\tilde{s})\gets \|(\mathbf{w}-\hat{\mathbf{w}}(\tilde{s}))\odot\sqrt{\mathbf{h}_i}\|_2^2$
    \EndFor
    \State $\tilde{s}^{*}\gets \arg\min_{\tilde{s}\in\mathcal{S}} E(\tilde{s})$
    \State Store $\hat{\mathbf{w}}_{r,i}\gets \hat{\mathbf{w}}(\tilde{s}^{*})$
\EndFor
\State \Return $\hat{\mathbf{W}}$
\end{algorithmic}
\end{algorithm}

\begin{algorithm}[t]
\caption{Combining 4over6 with H-Scale}
\label{alg:four_over_six_hscale}
\begin{algorithmic}[1]
\Require Row-group $\mathbf{w}_{r,i}$, Hessian weights $\mathbf{h}_i$, upward budget $U$, candidate budget $K$, global scale $s_{\mathrm{g}}$, 4over6 anchors $\mathcal{C}=\{4,6\}$
\Ensure Quantized row-group $\hat{\mathbf{w}}_{r,i}$
\State $m\gets \|\mathbf{w}_{r,i}\|_{\infty}$
\ForAll{$c\in\mathcal{C}$}
    \State $s_{c}\gets m/c$
    \State $\ell_{c}\gets Q_{\mathrm{fp8}}(s_{c}/s_{\mathrm{g}})$, \quad $\tilde{s}_{c}\gets s_{\mathrm{g}}\ell_{c}$
    \State $\hat{\mathbf{w}}_{c}\gets Q_{\mathrm{fp4}}(\mathbf{w}_{r,i}/\tilde{s}_{c})\,\tilde{s}_{c}$
    \State $L_{\mathrm{4over6}}(c)\gets \|\mathbf{w}_{r,i}-\hat{\mathbf{w}}_{c}\|_2^2$
\EndFor
\State $c^{\mathrm{init}}\gets \arg\min_{c\in\mathcal{C}} L_{\mathrm{4over6}}(c)$
\State $s^{\mathrm{init}}\gets \tilde{s}_{c^{\mathrm{init}}}$
\State \textcolor{red}{$\hat{\mathbf{w}}_{r,i}\gets \mathrm{HScaleGroupQuantize}(\mathbf{w}_{r,i}, \mathbf{h}_i, s^{\mathrm{init}}, U, K, s_{\mathrm{g}})$}
\State \Return the refined group $\hat{\mathbf{w}}_{r,i}$
\end{algorithmic}
\end{algorithm}

\begin{algorithm}[t]
\caption{Combining GPTQ with H-Scale}
\label{alg:gptq_hscale}
\begin{algorithmic}[1]
\Require Weight matrix $\mathbf{W}$, GPTQ Hessian $\mathbf{H}$, diagonal proxy $\mathbf{h}$, block size $B$, upward budget $U$, candidate budget $K$
\Ensure Quantized weight matrix $\hat{\mathbf{W}}$
\State Form the damped inverse-Cholesky factor $\mathbf{H}^{-1/2}$ as in GPTQ
\State Compute the layer-level NVFP4 global scale $s_{\mathrm{g}}$
\State Initialize $\hat{\mathbf{W}}\gets \mathbf{0}$
\For{$i=1,1+B,\ldots,n$}
    \State $\mathbf{W}_{i}\gets \mathbf{W}_{:,i:i+B-1}$, $\mathbf{H}_{i}\gets \mathbf{H}^{-1/2}_{i:i+B-1,i:i+B-1}$
    \State \textcolor{red}{$\mathbf{h}_{i}\gets \mathbf{h}_{i:i+B-1}$}
    \State \textcolor{red}{$\mathbf{s}^{\mathrm{init}}_{i}\gets$ baseline-provided effective NVFP4 scales for $\mathbf{W}_{i}$}
    \State \textcolor{red}{$\hat{\mathbf{W}}_{i}\gets \mathrm{HScaleGroupQuantize}(\mathbf{W}_{i}, \mathbf{h}_{i}, \mathbf{s}^{\mathrm{init}}_{i}, U, K, s_{\mathrm{g}})$}
    \ForAll{local columns $j$ in the block}
        \State $\mathbf{e}_{j}\gets (\mathbf{W}_{i,j}-\hat{\mathbf{W}}_{i,j}) / \mathbf{H}_{i,jj}$
    \EndFor
    \State $\hat{\mathbf{W}}_{:,i:i+B-1}\gets \hat{\mathbf{W}}_{i}$
    \State Update the remaining weights using the standard GPTQ error propagation:
    \[
        \mathbf{W}_{:,i+B:n}
        \gets
        \mathbf{W}_{:,i+B:n}
        -
        \mathbf{E}_{i}\mathbf{H}^{-1/2}_{i:i+B-1,i+B:n}.
    \]
\EndFor
\State \Return $\hat{\mathbf{W}}$
\end{algorithmic}
\end{algorithm}


\section{Evaluation Protocol Details}
\label{app:evaluation_protocol}

\textbf{Benchmark suites.} For compact table headers, we abbreviate MMLU-Redux as MMLU-R and LiveCodeBench as LCBench. Qwen3-30A3-Thinking uses the five-task suite in Table~\ref{tab:qwen30a3}. Qwen3-30A3-Instruct and Qwen3-4B-Instruct~\citep{yang2025qwen3technicalreport} use the eight-task suite in Tables~\ref{tab:qwen30a3_instruct} and~\ref{tab:qwen3-4b}. For LLaMA-3.1-8B-Instruct~\citep{grattafiori2024llama3herd}, we report a compact seven-task suite and omit tasks where the BF16 baseline is near random or zero, since they do not provide meaningful quantization comparisons.

\textbf{Decoding protocol.} Generation uses temperature $0.7$, top-$p{=}0.8$, top-$k{=}20$, repetition penalty $1.0$, and presence penalty $1.5$. LiveCodeBench uses beam size $10$, and AIME uses beam size $16$. Because stochastic decoding and benchmark sampling can introduce run-to-run fluctuation, we evaluate each method three times under this protocol and report the mean score. We do not attach statistical significance to small deltas or to cases where a quantized variant slightly exceeds BF16. We use the fixed protocol to keep paired comparisons controlled across methods.

\section{Run-to-Run Variability of the Main Results}
\label{app:variability}

Following the protocol of Appendix~\ref{app:evaluation_protocol}, every configuration is evaluated three times and the main-text tables report the per-run mean. This section reports the matching dispersion for the two Qwen3-30A3 suites: Table~\ref{tab:qwen30a3_std} for Table~\ref{tab:qwen30a3} and Table~\ref{tab:qwen30a3_instruct_std} for Table~\ref{tab:qwen30a3_instruct}. Run $k$ of a given configuration is a single end-to-end evaluation of all tasks, so the \textbf{Avg.} standard deviation is taken over the three per-run averages rather than propagated from the per-task columns.

\begin{table}[H]
\centering
\small
\caption{Run-to-run sample standard deviation (three runs, $n{=}3$, ddof${=}1$) for every entry of Table~\ref{tab:qwen30a3} (Qwen3-30A3-Thinking, five-task suite). The \textbf{Avg.} column is the standard deviation of the three per-run averages, not a propagation of the per-task values.}
\label{tab:qwen30a3_std}
\begin{tabular*}{\textwidth}{@{\extracolsep{\fill}}lcccccc@{}}
\toprule
  \textbf{Method} & \textbf{Avg.} & \textbf{AIME24} & \textbf{AIME25} & \textbf{MMLU-R} & \textbf{LiveBench} & \textbf{LCBench} \\ \midrule
  BF16 Baseline & 0.10 & 0.38 & 0.29 & 0.43 & 0.38 & 0.43 \\ \midrule
  RTN & 0.08 & 0.08 & 0.36 & 0.32 & 0.23 & 0.40 \\
  \quad + H-Scale & 0.13 & 0.22 & 0.18 & 0.19 & 0.53 & 0.55 \\ \midrule
  4over6 & 0.22 & 0.37 & 0.65 & 0.79 & 0.20 & 0.42 \\
  \quad + H-Scale & 0.14 & 0.64 & 0.63 & 0.25 & 0.47 & 0.37 \\ \midrule
  ArcQuant & 0.14 & 0.21 & 0.45 & 0.19 & 0.47 & 0.11 \\
  \quad + H-Scale & 0.19 & 0.80 & 0.43 & 0.57 & 0.22 & 0.33 \\ \midrule
  GPTQ & 0.37 & 0.26 & 0.36 & 0.17 & 0.61 & 0.59 \\
  \quad + H-Scale & 0.10 & 0.14 & 0.46 & 0.28 & 0.16 & 0.44 \\ \midrule
  GPTAQ & 0.06 & 0.43 & 0.23 & 0.64 & 0.37 & 0.25 \\
  \quad + H-Scale & 0.25 & 0.13 & 0.41 & 0.42 & 0.64 & 0.49 \\ \midrule
  MR-GPTQ & 0.20 & 0.26 & 0.70 & 0.32 & 0.33 & 0.65 \\
  \quad + H-Scale & 0.08 & 0.41 & 0.35 & 0.51 & 0.40 & 0.71 \\ \bottomrule
\end{tabular*}
\end{table}

\begin{table}[H]
\centering
\small
\caption{Run-to-run sample standard deviation (three runs, $n{=}3$, ddof${=}1$) for every entry of Table~\ref{tab:qwen30a3_instruct} (Qwen3-30A3-Instruct, eight-task suite). The \textbf{Avg.} column is the standard deviation of the three per-run averages, not a propagation of the per-task values.}
\label{tab:qwen30a3_instruct_std}
\setlength{\tabcolsep}{4.5pt}
\begin{tabular*}{\textwidth}{@{\extracolsep{\fill}}lccccccccc@{}}
\toprule
  \textbf{Method} & \textbf{Avg.} & \textbf{C-Eval} & \textbf{LiveBench} & \textbf{MMLU-R} & \textbf{AIME25} & \textbf{LCBench} & \textbf{ARC-C} & \textbf{BBH} & \textbf{GPQA} \\ \midrule
  BF16 Baseline & 0.10 & 0.10 & 0.47 & 0.37 & 0.63 & 0.54 & 0.38 & 0.25 & 0.49 \\ \midrule
  RTN & 0.10 & 0.39 & 0.51 & 0.20 & 0.41 & 0.57 & 0.53 & 0.35 & 0.46 \\
  \quad + H-Scale & 0.27 & 0.40 & 0.51 & 0.54 & 0.30 & 0.28 & 0.41 & 1.00 & 0.57 \\ \midrule
  4over6 & 0.13 & 0.34 & 0.22 & 0.43 & 0.24 & 0.47 & 0.29 & 0.31 & 0.41 \\
  \quad + H-Scale & 0.25 & 0.09 & 0.48 & 0.53 & 0.42 & 0.57 & 0.59 & 0.22 & 0.19 \\ \midrule
  ArcQuant & 0.17 & 0.35 & 0.93 & 0.23 & 0.24 & 0.18 & 0.47 & 0.29 & 0.14 \\
  \quad + H-Scale & 0.11 & 0.71 & 0.57 & 0.23 & 0.38 & 0.55 & 0.46 & 0.39 & 0.22 \\ \midrule
  GPTQ & 0.13 & 0.17 & 0.24 & 0.40 & 0.48 & 0.33 & 0.07 & 0.67 & 0.30 \\
  \quad + H-Scale & 0.14 & 0.78 & 0.42 & 0.28 & 0.51 & 0.29 & 0.62 & 0.86 & 0.25 \\ \midrule
  GPTAQ & 0.05 & 0.33 & 0.29 & 0.46 & 0.23 & 0.46 & 0.58 & 0.54 & 0.29 \\
  \quad + H-Scale & 0.16 & 0.47 & 0.31 & 0.18 & 0.17 & 0.42 & 0.27 & 0.34 & 0.42 \\ \midrule
  MR-GPTQ & 0.18 & 0.41 & 0.78 & 0.56 & 0.36 & 0.22 & 0.55 & 0.19 & 0.14 \\
  \quad + H-Scale & 0.12 & 0.01 & 0.35 & 0.63 & 0.38 & 0.20 & 0.59 & 0.54 & 0.35 \\ \bottomrule
\end{tabular*}
\end{table}

Per-task standard deviations range from $0.06$ to $1.00$ points with a median of $0.39$, and at three runs no benchmark in either suite is systematically tighter than the others. Averaging over the suite cancels most of this jitter: the \textbf{Avg.} standard deviation never exceeds $0.37$ points (median $0.14$), the shrinkage expected from averaging five and eight tasks respectively. Those standard-deviation results demonstrate the effectiveness of our method: the suite-level fluctuation is small relative to the reported average gains, confirming that H-Scale improves overall model performance rather than reflecting run-to-run noise.

\pagebreak
\section{Hessian Ablation on Qwen3-30A3-Thinking}
\label{app:hessian_ablation}

Table~\ref{tab:ablation_hessian} complements the ablation in the main text by repeating the same comparison on the Qwen3-30A3-Thinking benchmark suite. Scale-only optimization can recover part of the benefit over the raw baseline, but the Hessian-weighted objective gives the strongest average score for both ArcQuant and 4over6. This supports the interpretation that the gain is not merely from shrinking or retuning scales, but from weighting scale errors by their calibration-output sensitivity.

\begin{table}[H]
\centering
\small
\caption{Ablation on Qwen3-30A3-Thinking: scale with vs.\ without Hessian. ``Scale-only'' uses uniform unit weights in Eq.~\eqref{eq:scale_objective}.}
\label{tab:ablation_hessian}
\begin{tabular*}{\textwidth}{@{\extracolsep{\fill}}llcccccc@{}}
\toprule
\textbf{Baseline} & \textbf{Variant} & \textbf{Avg.} & \textbf{AIME24} & \textbf{AIME25} & \textbf{MMLU-R} & \textbf{LiveBench} & \textbf{LCBench} \\ \midrule
4over6 & Baseline & 80.32 & \textbf{87.82} & 77.04 & \textbf{90.83} & 75.34 & \textbf{70.56} \\
\quad & Scale-only (w/o $\mathbf{H}$) & 80.05 & 87.71 & 76.67 & 90.39 & \textbf{76.47} & 69.03 \\
\quad & + H-Scale & \textbf{80.35} & 87.44 & \textbf{77.84} & 90.72 & 75.53 & 70.20 \\ \midrule
ArcQuant & Baseline & 79.73 & 87.91 & 74.53 & 90.42 & \textbf{75.74} & \textbf{70.04} \\
\quad & Scale-only (w/o $\mathbf{H}$) & 79.68 & 87.08 & 75.42 & 90.78 & \textbf{75.74} & 69.39 \\
\quad & + H-Scale & \textbf{80.46} & \textbf{89.03} & \textbf{77.78} & \textbf{90.89} & 74.65 & 69.97 \\ \bottomrule
\end{tabular*}
\end{table}

\section{Calibration-Set Size}
\label{app:calib_size}

To further verify the robustness of H-Scale, we reduce the calibration set from 4096 to 128 samples while keeping all other settings unchanged. Table~\ref{tab:calib128} reports the result on Qwen3-30A3-Instruct: even with this much smaller budget, H-Scale still improves over the corresponding NVFP4 baseline.

\begin{table}[H]
\centering
\small
\caption{Qwen3-30A3-Instruct with a 128-sample calibration set.}
\label{tab:calib128}
\setlength{\tabcolsep}{4.5pt}
\begin{tabular*}{\textwidth}{@{\extracolsep{\fill}}lccccccccc@{}}
\toprule
\textbf{Method} & \textbf{Avg.} & \textbf{C-Eval} & \textbf{LiveBench} & \textbf{MMLU-R} & \textbf{AIME25} & \textbf{LCBench} & \textbf{ARC-C} & \textbf{BBH} & \textbf{GPQA} \\ \midrule
  BF16 Baseline & 65.58 & 63.43 & 69.27 & 89.11 & 59.90 & 48.42 & 69.05 & 70.50 & 54.99 \\ \midrule
  GPTQ & 62.70 & 62.08 & \textbf{66.10} & 87.87 & \textbf{56.25} & 47.74 & 67.32 & 64.75 & 49.49 \\
  \quad + H-Scale & \textbf{63.32} & \textbf{62.97} & 65.56 & \textbf{87.90} & 55.42 & \textbf{48.14} & \textbf{67.89} & \textbf{66.15} & \textbf{52.53} \\ \midrule
  GPTAQ & 63.44 & 62.75 & 66.31 & \textbf{87.76} & 54.33 & 48.21 & 67.41 & 66.20 & 54.55 \\
  \quad + H-Scale & \textbf{64.47} & \textbf{63.84} & \textbf{67.41} & 87.47 & \textbf{58.17} & 48.21 & \textbf{67.92} & \textbf{67.72} & \textbf{55.05} \\ \bottomrule
\end{tabular*}
\end{table}

\end{document}